\documentclass[runningheads]{llncs}
\usepackage{subcaption}
\usepackage{graphicx}
\usepackage{amsmath,amssymb} 
\usepackage{mathtools}
\usepackage[nocompress]{cite}

\DeclarePairedDelimiter{\ceil}{\lceil}{\rceil}

\usepackage{color}
\usepackage[nolist]{acronym}
\usepackage[width=122mm,left=12mm,paperwidth=146mm,height=193mm,top=12mm,paperheight=217mm]{geometry}

\newif\ifdotpgf\dotpgffalse

\ifdotpgf
\usepackage{pgfplots}
\usepgfplotslibrary{external}
\pgfrealjobname{0587}
\else
\long\def\beginpgfgraphicnamed#1#2\endpgfgraphicnamed{\includegraphics{#1}}
\fi

\begin{acronym}
    \acro{CNN} {Convolutional Neural Network}
    \acro{HOG} {Histogram of Oriented Gradients}
    \acro{ReLU} {Rectified Linear Unit}
    \acro{LCN} {Local Contrast Normalization}
    \acro{NMS} {Non Maxima Suppression}
    \acro{SGD} {Stochastic Gradient Descent}
    \acro{FPPI} {False Positives Per Image}
    \acro{DPM} {Deformable Parts Model}
    \acro{LFPW} {Labeled Face Parts in the Wild}
    \acro{AFLW} {Annotated Facial Landmarks in the Wild}
    \acro{AFW} {Annotated Faces in the Wild}
    \acro{FDDB} {Face Detection Data Set and Benchmark}
    \acro{SSE} {Sum of Squares Error}
    \acro{ACF} {Aggregate Channel Features}
    \acro{AP} {Average Precision}
    \acro{COFW} {Caltech Occluded Faces in the Wild}
\end{acronym}

\begin{document}


\pagestyle{headings}
\mainmatter

\title{Grid Loss: Detecting Occluded Faces} 

\titlerunning{Grid Loss: Detecting Occluded Faces}

\authorrunning{M. Opitz, G. Waltner, G. Poier, H. Possegger, and H. Bischof}

\author{Michael Opitz, Georg Waltner, Georg Poier, Horst Possegger, Horst Bischof}
\institute{Institute for Computer Graphics and Vision \\ 
           Graz University of Technology \\ 
           \email{\{michael.opitz,waltner,poier,possegger,bischof\}@icg.tugraz.at}}

\maketitle

\begin{abstract} 
Detection of partially occluded objects is a challenging computer vision
problem.
Standard Convolutional Neural Network (CNN) detectors fail if parts of the detection window are occluded,
since not every sub-part of the window is discriminative on its own.
To address this issue, we propose a novel loss layer for CNNs, 
named \emph{grid loss}, which minimizes the error rate on sub-blocks of a convolution layer
independently rather than over the whole feature map. This results in 
parts being more discriminative on their own, enabling the detector to recover
if the detection window is partially occluded.
By mapping our loss layer back to a regular 
fully connected layer, no additional computational cost is incurred at
runtime compared to standard CNNs. We demonstrate our method for face detection
on several public face detection benchmarks and show that our method outperforms
regular CNNs, is suitable for realtime applications and achieves state-of-the-art performance.
\end{abstract}

\begin{figure}[!h]
    \centering
    \begin{subfigure}[t]{0.40\textwidth}
        \centering
        \includegraphics[width=0.45\textwidth]{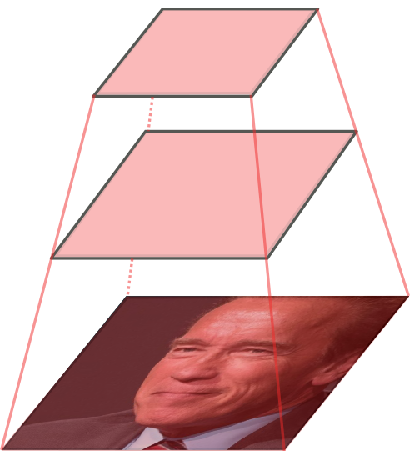}
        \includegraphics[width=0.45\textwidth]{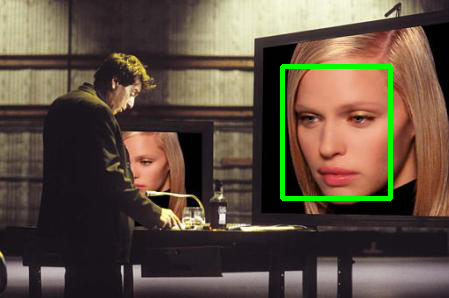}
        \caption{Global loss layer.}
        \label{fig:teaser-image:a}
    \end{subfigure}
    \hspace{0.1cm}
    \begin{subfigure}[t]{0.40\textwidth}
        \centering
            \includegraphics[width=0.45\textwidth]{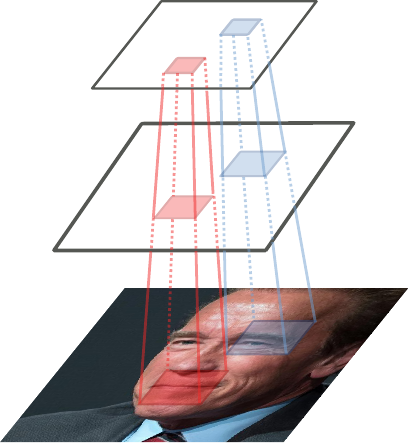}
            \includegraphics[width=0.45\textwidth]{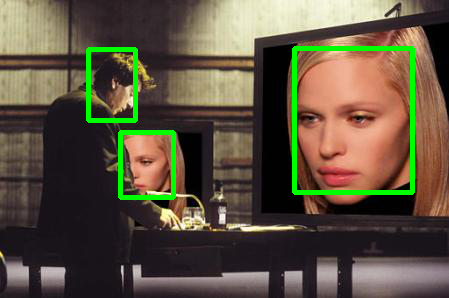}
        \caption{Our grid loss layer.}
        \label{fig:teaser-image:b}
    \end{subfigure}
    \caption{Schematic overview of~\subref{fig:teaser-image:a} standard global loss and~\subref{fig:teaser-image:b} the proposed grid loss with an illustrative example on FDDB.}
    \label{fig:teaser-image}
\end{figure}

\section{Introduction}
%
We focus on single-class object detection and
in particular address the problem of face detection.
Several applications for face detection, such as surveillance or robotics, impose realtime requirements and rely
on detectors which are fast, accurate and have low memory overhead.
Traditionally, the most prominent approaches have been based on boosting~\cite{Benenson2013Cvpr, DollarPAMI14pyramids, 
Mathias2014Eccv, schulter14a, Viola2004, Zhang2015Cvpr,yang2015convolutional}
and \acp{DPM}~\cite{felzenszwalb2010object, Mathias2014Eccv}.
More recently, following the success of deep learning for computer vision, e.g. \cite{krizhevsky2012imagenet},
methods based on \acp{CNN} have been applied to single-class object 
detection tasks, e.g.~\cite{farfade2015multi, Hosang2015Cvpr, CNNCascade2015, sermanet2013pedestrian}.

One of the most challenging problems in the context of object detection is handling partial occlusions.
Since the occluder might have arbitrary appearance, occluded objects have significant intra-class variation.
Therefore, collecting 
large datasets capturing the huge variability of occluded objects, which is required for training large \acp{CNN}, is expensive. The main question we address
in this paper is: How can we train a \ac{CNN} to detect occluded objects?


In standard \acp{CNN} not every sub-part of the detection template is 
discriminative alone (i.e.\ able to distinguish faces from background), 
resulting in missed faces if parts of the detection template are occluded.
Our main contribution is to address this issue by introducing a novel loss layer for 
\acp{CNN}, named \emph{grid loss}, which is illustrated in Fig.~\ref{fig:teaser-image}.  This layer divides
the convolution layer into spatial blocks and optimizes the hinge loss on each
of these blocks separately. This results in several independent detectors which
are discriminative on their own. If one part of the window is occluded, only a
subset of these detectors gets confused, whereas the remaining ones will still
make correct predictions.


By requiring parts to be already discriminative on their own, we encourage
the \ac{CNN} to learn features suitable for classifying parts of an object. 
If we would train a loss over the full face, the \ac{CNN} might
solve this classification problem by just learning features which detect a subset of discriminative regions, e.g. 
eyes. We divide our window into sub-parts and some of these
parts do not contain such highly prototypical regions. Thus, the \ac{CNN} has to also learn discriminative
representations for other parts corresponding to e.g. nose or mouth. We find that \acp{CNN} trained with grid 
loss develop more diverse and independent features compared to \acp{CNN} trained with a regular loss.

After training we map our grid loss layer back to a regular fully connected layer. 
Hence, no additional runtime cost is incurred by our method. 

As we show in our experiments, grid loss significantly improves over 
using a regular linear layer on top of a convolution layer without imposing 
additional computational cost at runtime. We evaluate our
method on publicly available face detection datasets~\cite{fddbTech,
Yan2014790, ZhuRCVPR2012} and show that it compares favorably to state-of-the-art methods.  
Additionally, we present a detailed parameter evaluation 
providing further insights into our method, which shows that grid loss especially benefits detection of occluded faces and 
reduces overfitting by efficiently combining several spatially independent detectors. 

\section{Related Work}
\label{sec:related-work}
Since there is a multitude of work in the area of face detection, a 
complete discussion of all papers is out of scope of this work. Hence, we focus
our discussion only on seminal work and closely related approaches in the field and refer to \cite{zafeiriou2015survey} for a more complete survey.

A seminal work is the method of
Viola and Jones~\cite{Viola2004}. They propose a realtime detector using a cascade 
of simple decision stumps. These classifiers are based on 
area-difference features computed over differently sized rectangles.
To accelerate feature computation, they employ integral images for computing
rectangular areas in constant time, independent of the rectangle size.

Modern boosting based detectors use linear classifiers on SURF based features~\cite{li2013learning}, 
exemplars~\cite{li2014efficient}, and leverage landmark
information with shape-indexed features for classification~\cite{chen2014joint}.  
Other boosting based detectors compute integral images
on oriented gradient features as well as LUV channels and use shallow boosted
decision trees~\cite{Mathias2014Eccv} or constrain the features on the
feature channels to be block sized~\cite{yang2014aggregate}. 
Additionally, \cite{yang2015convolutional} proposes \ac{CNN} features for the boosting framework.

Another family of detectors are \ac{DPM}~\cite{felzenszwalb2010object} based detectors, which learn
root and part templates. The responses of these templates are combined with a deformation 
model to compute a confidence score. Extensions to \acp{DPM} have been proposed which 
handle occlusions~\cite{ghiasi2014occlusion}, improve runtime speed~\cite{yan2014} and leverage
manually annotated part positions in a tree structure~\cite{ZhuRCVPR2012}. 

Further, there are complimentary approaches improving existing detectors by domain 
adaption techniques~\cite{li2013probabilistic}; and exemplar based methods using 
retrieval techniques to detect and align faces~\cite{shen2013detecting, kumar2015}.

Recently, \acp{CNN} became increasingly popular due to their success in
recognition and detection problems, e.g.~\cite{girshick2014rich,
krizhevsky2012imagenet}.  They successively apply convolution filters followed
by non-linear activation functions. Early work in this area applies a small
number of convolution filters followed by sum or average pooling on the
image~\cite{FaceFinder2004, rowley1998neural, vaillant1994original}. More
recent work leverages a larger number of filters which are pre-trained on large datasets, e.g.
ILSVRC \cite{ILSVRC15}, and fine-tuned on face datasets.  These
approaches are capable of detecting faces in multiple orientations and poses,
e.g.~\cite{farfade2015multi}. Furthermore,~\cite{CNNCascade2015} uses a
coarse-to-fine neural network cascade to efficiently detect faces in realtime.
Successive networks in the cascade
have a larger number of parameters and use previous features of the cascade as
inputs. \cite{yang2015facial} propose a large dataset with attribute annotated
faces to learn 5 face attribute \acp{CNN} for predicting hair, eye, nose, mouth
and beard attributes (e.g.\ black hair vs.\ blond hair vs.\ bald hair).
Classifier responses are used to re-rank object proposals, which are then
classified by a CNN as face vs.\ non-face.

In contrast to recent \ac{CNN} based approaches for face detection \cite{yang2015facial, farfade2015multi, CNNCascade2015}, we exploit
the benefits of part-based models with our grid loss layer by efficiently
combining several spatially independent networks to improve detection performance
and increase robustness to partial occlusions. Compared to \cite{yang2015facial}, our method does not require
additional face-specific attribute annotations and is more generally applicable to other object detection problems.
Furthermore, our method is suitable for realtime applications.

\section{Grid Loss for \acp{CNN}}
\label{sec:methodology}
%
%
\begin{figure*}[t!]
    \centering
    \includegraphics[width=0.9\textwidth]{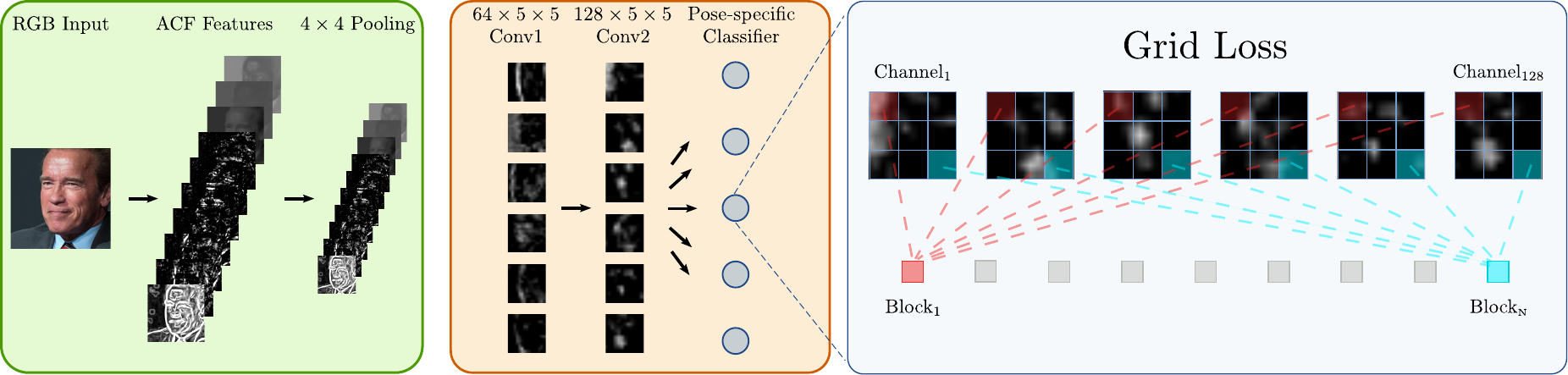}
    \caption{Overview of our method: our detection CNN builds upon \acf{ACF}~\cite{DollarPAMI14pyramids}. For each window, after pooling, we apply successive convolution
    filters to the input channels. To distinguish faces from non-faces we use pose-specific classifiers. Instead of minimizing
    the loss over the last full convolution map, we divide the map into small blocks and minimize a loss function on each of these blocks independently. We train our CNN end-to-end with backpropagation.}
    \label{fig:method-overview}
\end{figure*}

We design the architecture of our detector based on the following key requirements for
holistic detectors: We want to achieve realtime performance to process video-stream data and 
achieve state-of-the-art accuracy.
To this end, we use the network architecture as illustrated in Fig.~\ref{fig:method-overview}. 
Our method detects faces using a sliding window, similar to \cite{sermanet-iclr-14}.
We apply two convolution layers on top of the input features as detailed in Sec.~\ref{sec:architecture}.
In Sec.~\ref{sec:grid-loss-layer}, we introduce our grid loss layer to obtain highly accurate part-based pose-specific classifiers.
Finally, in Sec.~\ref{sec:regression} we propose a regressor to refine face positions and skip several intermediate octave levels
to improve runtime performance even further.



\subsection{Neural Network Architecture}
\label{sec:architecture}
The architecture of our \ac{CNN}
consists of two $5 \times 5$ convolution layers (see Fig.~\ref{fig:method-overview}). 
Each convolution layer is followed by a
\ac{ReLU} activation. To normalize responses across layers, we use a \ac{LCN} layer in between the two
convolution layers. Further, we apply a small amount of dropout~\cite{srivastava2014dropout} of $0.1$ after the last convolution layer. 
We initialize the weights randomly with a Gaussian of zero mean and $0.01$ standard deviation. Each unit in the output layer corresponds to a specific face pose, which
is trained discriminatively against the background class. We define the
final confidence for a detection window as the maximum confidence over
all output layer units.


In contrast to other \ac{CNN} detectors, mainly for speed reasons, we use \acf{ACF}~\cite{DollarPAMI14pyramids} as low-level
inputs to our network.
For face detection we subsample the \ac{ACF} pyramid 
by a factor of~4, reducing the computational cost of the
successive convolution layers.

At runtime, we apply the \ac{CNN} detector in a sliding window fashion densely over the feature pyramid at several
scales. After detection, we perform \ac{NMS} of two bounding boxes
$B_a$ and $B_b$ using the overlap score $o_{\textrm{\tiny NMS}}(B_{a}, B_{b}) = \frac{\left| B_{a}
\cap B_{b} \right|}{\min(\left| B_a \right|, \left| B_b \right|)}$, where
$\left| B_a \cap B_b \right|$ denotes the area of intersection of the two
boxes and $\min(\left| B_a \right|, \left| B_b \right| )$ denotes the
minimum area of the two boxes.  Boxes are suppressed if their overlap
exceeds $0.3$, following~\cite{Mathias2014Eccv}.


\begin{figure}[t]
    \centering
    \begin{subfigure}[t]{0.45\textwidth}
        \includegraphics[width=\textwidth]{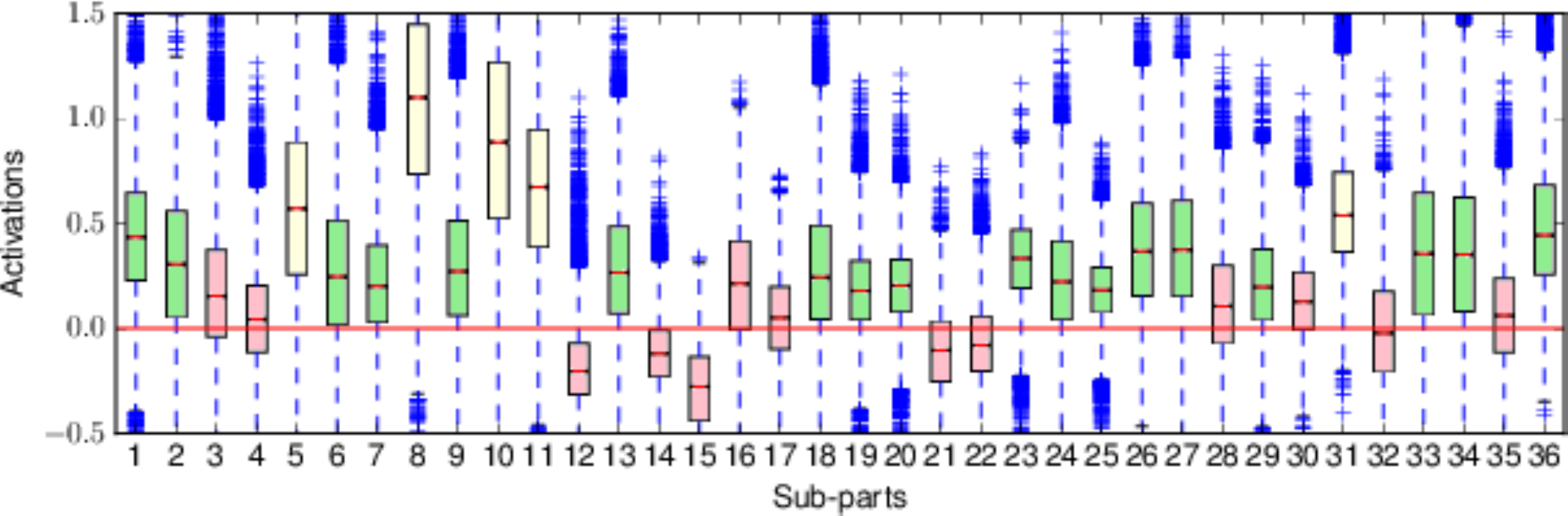}
        \caption{Regular loss}
        \label{fig:boxplot-regular}
    \end{subfigure}
    \begin{subfigure}[t]{0.45\textwidth}
        \includegraphics[width=\textwidth]{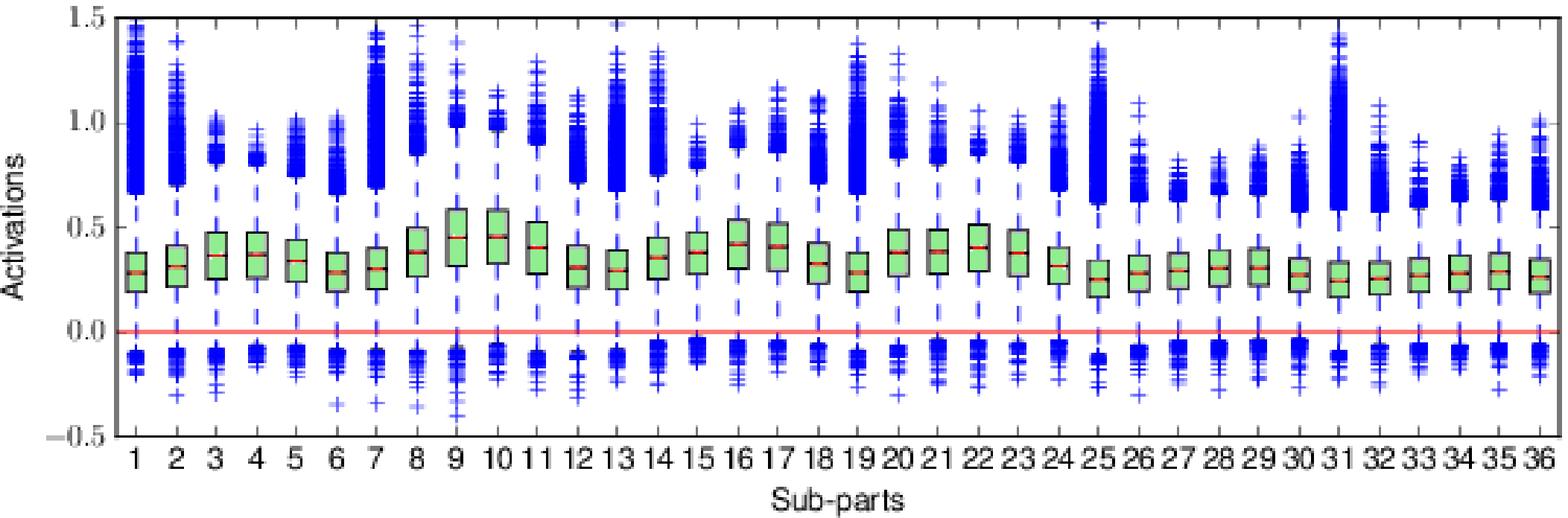}
        \caption{Grid loss}
        \label{fig:boxplot-grid}
    \end{subfigure}

    \caption{Boxplot of $2\times2$ part activations on the positive training set (i.e. by dividing the detection template into non-overlapping parts, as in Fig. \ref{fig:method-overview}). 
        Activations trained by regular loss
        functions can have parts with negative median response. 
        We mark parts whose 25\% percentile is smaller than $0$ (red) and parts which have significant positive
        median activations compared to other parts (yellow).}
    \label{fig:boxplot}
\end{figure}

\subsection{Grid Loss Layer}
\label{sec:grid-loss-layer}
\ac{CNN} detection templates can have non-discriminative sub-parts, which
produce negative median responses over the positive training set (see Fig. \ref{fig:boxplot-regular}). To achieve an overall positive prediction
for a given positive training sample, they heavily rely on certain sub-parts of a feature map to make a strong positive prediction. 
However, if these parts are occluded, 
the prediction of the detector is negatively influenced.
To tackle this problem, we propose to divide the convolution layers
into small $n \times n$ blocks and optimize the hinge loss for each of these blocks separately. This results in a
detector where sub-parts are discriminative (see Fig \ref{fig:boxplot-grid}). If a part of an input face
is occluded, a subset of these detectors will still have non-occluded face parts as inputs.
More formally, let $\boldsymbol{x}$ denote a vectorized $f \times r \times c$
dimensional tensor which represents the last convolution layer map, where $f$ denotes the number of filters,
$r$ denotes the number of rows and $c$ the number of columns of the feature map. 
We divide $\boldsymbol{x}$ into small
$f \times n \times n$ non-overlapping blocks $\boldsymbol{f}_i$, ${i = 1 \ldots N}$, with 
${N =\ceil{\frac{r}{n}} \cdot \ceil{\frac{c}{n}}}$.
To train our layer, we use the hinge loss
\begin{align}
    l(\boldsymbol{\theta}) = \sum_{i=1}^{N} \max(0, m - y \cdot (\boldsymbol{w}_i^{\top} \boldsymbol{f}_i + b_i)), 
    \label{eqn:block-loss}
\end{align}
where $\boldsymbol{\theta} = [\boldsymbol{w}_1, \boldsymbol{w}_2, \ldots,
\boldsymbol{w}_N, b_1, b_2, \ldots, b_N]$, $m$ is the margin, $y \in \{-1, 1\}$
denotes the class label, $\boldsymbol{w}_i$ and $b_i$ are the weight
vector and bias for block $i$, respectively.  In all our experiments we set $m$ to
$\frac{1}{N}$, since each of the $N$ classifiers is responsible to push a given
sample by $\frac{1}{N}$ farther away from the separating hyperplane.

Since some of the part classifiers might correspond to less discriminative face parts, we
need to weight the outputs of different independent detectors correctly. Therefore,
we combine this local per-block loss with a global hinge loss
which shares parameters with the local classifiers. We concatenate the
parameters $\boldsymbol{w} = [\boldsymbol{w}_1, \boldsymbol{w}_2, \ldots, \boldsymbol{w}_N]$ and set $b = \sum_i b_i$. Our final
loss function is defined as
\begin{equation}
    l(\boldsymbol{\theta}) =  \max(0, 1 - y \cdot (\boldsymbol{w}^{\top} \boldsymbol{x} + b)) + 
                              \lambda \cdot \sum_{i=1}^{N} \max(0, m - y \cdot (\boldsymbol{w}_i^{\top} \boldsymbol{f}_i + b_i)),
\end{equation}
where $\lambda$ weights the individual part detectors vs. the holistic detector and is empirically set to $1$ in our
experiments (see Sec.~\ref{sec:eval-grid-loss-weighting}). To
optimize this loss we use \ac{SGD} with momentum. Since the weights $\boldsymbol{w}$ are shared 
between the global and local classifiers and $b$ is a sum of existing parameters, the number of additional parameters 
is only $N - 1$ compared to a
regular classification layer. However, at runtime no additional computational cost
occurs, since we concatenate the local weight vectors to form a global weight vector
and sum the local biases to obtain a global bias.

During training, the holistic loss backpropagates an error for misclassified samples to the hidden layers.
Also, if certain parts are misclassifying a given sample, the part loss backpropagates an additional error 
signal to the hidden layers. However, for part detectors which are already
discriminative enough to classify this sample correctly, no additional part error 
signal is backpropagated.
In this way error signals of less discriminative parts are strengthened during training, encouraging the
\ac{CNN} to focus on making weak parts stronger rather than strengthening already discriminative parts (see Fig.~\ref{fig:boxplot-grid}). 
This can also be observed when a sample is correctly classified by the holistic detector, but is misclassified by 
some part detectors. In this case only an error signal from the part classifiers is backpropagated, resulting in the
part detectors becoming more discriminative.
By training a \ac{CNN} this way, the influence of several strong distinguished parts decreases, since they cannot 
backpropagate as many errors as non-discriminative parts, resulting in a more
uniform activation pattern across parts, as seen in Fig. \ref{fig:boxplot}.
With more uniform activations, even if some parts fail due to occlusions, the detector can recover.
We experimentally confirm robustness to occlusions of our method in Sec.~\ref{sec:occlusion-robustness}.  
%
\subsubsection{Regularization Effect.}
Good features are highly discriminative and decorrelated, so that they are complementary if they are composed.
Another benefit of grid loss is that it reduces correlation of feature maps
compared to standard loss layers, which we experimentally show in Sec.~\ref{sec:grid-loss-overfitting}. 
We accredit this to the fact that the loss encourages parts to be discriminative. For a holistic detector 
a \ac{CNN} might rely on a few mid-level features to classify a window as face or background. In contrast to that, 
with grid loss the \ac{CNN} has to learn mid-level features which can distinguish each face part from the background,
resulting in a more diverse set of mid-level features. More diverse features result in activations which are decorrelated. 
Another interpretation of our method is, that we perform efficient model averaging of several part-based detectors with a shared
feature representation, which 
reduces overfitting.
We show in 
Sec.~\ref{sec:eval-grid-loss-training-size} that with a smaller training set size the 
performance difference to standard loss functions increases compared to grid loss.
%
\subsubsection{Deeply Supervised Nets.}
The output layer of a neural network has a higher chance of discriminating
between background and foreground windows if its features are discriminative.
Previous works~\cite{Szegedy2014,li2014efficient} improve the discriminativeness of their feature
layers for object classification by applying a softmax or hinge loss on top of their hidden layers.
Inspired by this success we replace the standard loss with our grid loss and apply it on top of our hidden layers.
As our experiments show (Sec.~\ref{sec:eval-grid-loss}), this further improves the performance without sacrificing 
speed, since
these auxiliary loss layers are removed in the classification step.

\subsection{Refinement of Detection Windows}
\label{sec:regression}
Sliding window detectors can make mislocalization errors, causing high confidence 
predictions to miss the face by a small margin. This results in highly confident false positive predictions. To correct these errors, 
we apply a regressor to refine the location
of the face. 
Further, we empirically observe that our \ac{CNN} with the proposed grid loss is able to detect 
faces which are slightly smaller or bigger than the sliding window. 
Tree based detectors use an image pyramid with $8$ intermediate scales per octave. Applying several convolutions on 
top of all these scales is computationally expensive. Based on our observation, we propose to omit several of these intermediate scales and rely on the
regressor to refine the face location.
Details of this \ac{CNN} are provided in the supplementary material.

Evaluation protocols for face detection use the PASCAL VOC overlap criterion to
assess the performance. For two faces $F_a$ and $F_b$, the overlap $o_{\textrm{\tiny VOC}}$ is defined as
\begin{equation}
    \label{eqn:pascal}
    o_{\textrm{\tiny VOC}}(F_a, F_b) = \frac{\left| F_a \cap F_b \right|}{\left| F_a \cup F_b \right|},
\end{equation}
where $\left| F_a \cap F_b \right|$  denotes the intersection and $\left| F_a \cup F_b \right|$ denotes the
union of two face representations, i.e.\ ellipses or bounding boxes. 

For ellipse predictions, the parameters major
and minor axis length, center coordinates and orientation impact the PASCAL
overlap criteria differently. For example, a difference of 1 radiant in orientation changes
the overlap of two ellipses more than a change of 1 pixel in major axis length.
To account for these differences, we compare minimizing the standard \ac{SSE} error with maximizing the 
PASCAL overlap criteria in Equation~\eqref{eqn:pascal} directly. We
compute the gradient entries $g_i$, $i = 1, \ldots, 5$, of the loss function numerically by central differences:
\begin{equation}
    g_i(\boldsymbol{r}) \approx \frac{o_{\textrm{\tiny VOC}}(\boldsymbol{r} + \epsilon_i \cdot \boldsymbol{a}_i, \boldsymbol{y}) - o_{\textrm{\tiny VOC}}(\boldsymbol{r} - \epsilon_i \cdot \boldsymbol{a}_i, \boldsymbol{y})}{2 \cdot \epsilon_i},
\end{equation}
where $\boldsymbol{r}$ denotes the regressor predictions for the ellipse parameters, $\boldsymbol{y}$ denotes the 
ground truth parameters, $\boldsymbol{a}_i$ denotes the $i$-th standard basis vector where only the $i$-th 
entry is nonzero and set to $1$ and $\epsilon_i$ is the step size. Since the input size of
this network is $40\times40$ pixels, we use a patch size of $40\times40$ pixels
to rasterize both the ground truth ellipse and the predicted ellipse. Furthermore, we choose
$\epsilon_i$ big enough so that the rasterization changes at least by one pixel.
\section{Evaluation}
\label{sec:results}
We collect 15,106 samples from the \ac{AFLW}~\cite{koestinger11b} dataset to train our detector on $80
\times 80$ pixel windows in which $60 \times 60$ faces are visible. Similar 
to~\cite{Mathias2014Eccv}, we group faces into 5 discrete poses by yaw angle and
constrain faces to have pitch and roll between -22 and +22 degrees. Further
following~\cite{Mathias2014Eccv}, we create rotated versions of each pose by
rotating images by 35 degrees. We discard grayscale training images, since \acp{ACF} are color based. 
Finally, we mirror
faces and add them to the appropriate pose-group to augment the dataset.

We set the \ac{ACF} pre-smoothing radius to 1, the subsampling
factor to 4 and the post-smoothing parameter to 0.   Since we shrink the feature maps 
by a factor of 4, our \ac{CNN} is trained on $20 \times 20$ input patches
consisting of 10 channels.

For training we first randomly subsample 10,000 negative examples from the non-person
images of the PASCAL VOC dataset~\cite{pascal}. To estimate convergence of \ac{SGD} 
in training, we use 20\% of the data as validation set and the remaining 80\% as training set.
The detector is bootstrapped by collecting 10,000 negative patches in each bootstrapping iteration. After 3
iterations of bootstrapping, no hard negatives are detected.

Our regressor uses input patches of twice the size of our detector to capture
finer details of the face. Since no post-smoothing is used, we reuse the feature
pyramid of the detector and crop windows from one octave lower than they are
detected.

We evaluate our method on three challenging public datasets:
\ac{FDDB}~\cite{fddbTech},
\ac{AFW}~\cite{ZhuRCVPR2012}
and PASCAL Faces~\cite{Yan2014790}.  
\ac{FDDB} consists of 2,845 images with 5,171 faces and uses ellipse annotations.
PASCAL Faces is extracted from 851 PASCAL VOC images and has 1,635 faces and
\ac{AFW} consists of 205 images with 545 faces. Both \ac{AFW} and PASCAL Faces use bounding box annotations.

\subsection{Grid Loss Benefits}
\label{sec:eval-grid-loss}
%
To show the effectiveness of our grid loss layer we run
experiments on FDDB~\cite{fddbTech} using the neural network architecture
described in Sec.~\ref{sec:architecture} under the evaluation protocol described in \cite{fddbTech}.  For these experiments we do not
use our regressor to exclude its influence on the results and apply the network densely across all 8 intermediate
scales per octave (i.e.\ we do not perform layer skipping or location refinement).  We compare standard logistic loss, hinge loss and our
grid loss at a false positive count of 50, 100, 284 (which corresponds to $\approx$ 0.1 \ac{FPPI}) and 500 samples. 
Further, during training we apply grid loss to our hidden layers to improve the discriminativeness of our feature maps. 
In Table~\ref{tbl:fddb-eval-hinge} we see that our grid 
loss performs
significantly better than standard hinge or logistic loss, improving true positive rate by 3.2\% at 0.1 FPPI. 
Further, 
similar to the findings of~\cite{Szegedy2014, li2014efficient} our grid loss also 
benefits from auxiliary loss layers on top of hidden layers during training and additionally improves the true positive rate 
over the baseline by about 1\%.

\begin{table}[h]
    \centering
    \begin{subfigure}{0.4\textwidth} 

    \resizebox{1.0\textwidth}{!}{
        \beginpgfgraphicnamed{fddb-all-loss-functions}
        \input{images/fddb-all-loss-functions.pgf}
        \endpgfgraphicnamed
    }
    \end{subfigure}
    \begin{subfigure}{0.45\textwidth} 
    \small
    \begin{tabular}{l|cccc}
    \hline
    Method & 50 FP  & 100 FP  & 284 FP  & 500 FP  \\
    \hline
    L  & 0.776 & 0.795 & 0.817 & 0.824 \\
    H  & 0.758 & 0.786 & 0.819 & 0.831 \\
    \hline
    G+L & 0.803 & 0.827 & 0.851 & 0.859 \\
    G+H & 0.807 & 0.834 & 0.851 & 0.858 \\
    \hline
    G-h+L & \underline{0.809} & \underline{0.836} & \underline{0.862} & \underline{0.869}  \\
    G-h+H & \textbf{0.815} & \textbf{0.838} & \textbf{0.863} & \textbf{0.871}  \\
    \hline
    \end{tabular}
    \end{subfigure}
    \caption{True positive rates of logistic (L), hinge (H), grid + logistic (G+L), grid + hinge (G+H), grid hidden + hinge (G-h+H) and grid hidden + logistic (G-h+L) loss functions on FDDB at a false positive (FP) count of 50, 100, 284 and 500. \textbf{Best} and \underline{second best} results are highlighted.}
    \label{tbl:fddb-eval-hinge}
\end{table}

\subsection{Block Size}
\label{sec:eval-grid-loss-block-size}
To evaluate the performance of our layer with regard to the block size, we train several models with
different blocks of size $n = 2^{\{1,2,3,4\}}$ in the output and hidden layer. We constrain
the block size of the hidden layers to be the same as the block size of the output layers. Results
are shown in Table~\ref{tbl:block-size-comparison}. 
Our layer works best with small blocks of size 2 and degrades gracefully with larger blocks.
In particular, if the size is increased to 16 the method corresponds to a standard \ac{CNN} regularized with the method proposed
in~\cite{Szegedy2014, lee2014deeply} and thus, the grid loss layer does not show additional benefits.

\begin{table}[h]
    \centering
    \begin{subfigure}{0.4\textwidth}
    \resizebox{1.0\textwidth}{!}{
        \beginpgfgraphicnamed{fddb-box-size-comparison}
        \input{images/fddb-box-size-comparison.pgf}
        \endpgfgraphicnamed
    }
    \end{subfigure}
    \begin{subfigure}{0.45\textwidth}
    \centering
    \small
    \begin{tabular}{l|cccc}
    \hline
    Method  & 50 FP  & 100 FP  & 284 FP  & 500 FP \\
    \hline
    Block-2 & \textbf{0.815} & \textbf{0.838} & \textbf{0.863} & \textbf{0.871} \\
    Block-4 & \underline{0.812} & \underline{0.834} & \underline{0.852} & \underline{0.861}  \\
    Block-8 & 0.790 & 0.809 & 0.830 & 0.838 \\
    Block-16 & 0.803 & 0.816 & 0.834 & 0.843 \\
    \hline
    \end{tabular}

    \end{subfigure}

    \caption{Comparison of different block sizes on FDDB.}
    \label{tbl:block-size-comparison}
\end{table}

\begin{figure}[h!]
    \centering
    \resizebox{0.3\textwidth}{!}{
        \beginpgfgraphicnamed{fddb-weight-evaluation}
        \input{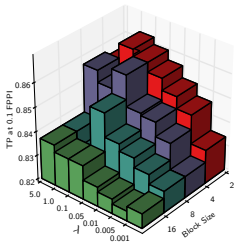}
        \endpgfgraphicnamed
    }
    \caption{Evaluation of the weighting parameter $\lambda$.}
    \label{fig:weighting-parameter-evaluation}
\end{figure}

\subsection{Weighting Parameter}
\label{sec:eval-grid-loss-weighting}
To evaluate the impact of the weighting parameter $\lambda$, we conduct
experiments comparing the true positive rate of our method at a false
positive count of 284 ($\approx$~0.1~\ac{FPPI}) with block sizes of
$2^{\{1,2,3,4\}}$ and ${\lambda = \{5, 1, 0.1, 0.05, 0.01, 0.005, 0.001\}}$.

Fig.~\ref{fig:weighting-parameter-evaluation} shows that our method
performs best with ${\lambda \approx 1}$ and smaller blocks of size 2 or 4. 
The performance of our method stays stable until $\lambda$ is varied more than
one order of magnitude. As $\lambda$ decreases, the network converges to the performance of a regular \ac{CNN} trained on hinge loss. 

\subsection{Robustness to Occlusions}
\label{sec:occlusion-robustness}
To show that grid loss helps to detect faces with occlusions,
we run an experiment on the \ac{COFW} dataset \cite{burgos2013robust}. The
original purpose of the \ac{COFW} dataset is to test facial landmark
localization under occlusions. It consists of 1,852 faces with
occlusion annotations for landmarks. We split the dataset into 329 heavily
occluded faces with $\ge 30\%$ of all landmarks occluded (COFW-HO) and 1,523
less occluded faces (COFW-LO). Since this dataset is proposed for landmark localization,
the images do not contain a large background variation.

For a fair evaluation, we measure the
\ac{FPPI} on \ac{FDDB}, which has a more realistic background variation for the
task of face detection. We report here the true positive rate on \ac{COFW} at $0.1$
\ac{FPPI} on \ac{FDDB}. This evalution ensures that the detectors achieve a low false positive
rate in a realistic detection setting and still detect occluded faces.

We evaluate both, the grid loss detector and the hinge loss detector
on this dataset. The
performance difference between these two detectors should increase on the occluded
subset of \ac{COFW}, since grid loss is beneficial for detecting occluded
faces. In Table~\ref{tbl:occlusion-robustness} we
indeed observe that the performance difference on the heavily occluded subset
significantly increases from 1.6\% to 7\% between the two detectors, demonstrating the favourable performance of 
grid loss for detecting occluded objects.

\begin{table}[h]
    \centering
    \begin{minipage}[t]{0.61\textwidth}
    \centering
    \begin{tabular}{l|cc}
        \hline
        Method & COFW-HO & COFW-LO \\
        \hline
        G &  \textbf{0.979} & \textbf{0.998} \\
        H &  0.909 & 0.982 \\
        \hline
    \end{tabular}
    \caption{True Positive Rate on COFW Heavily Occluded (COFW-HO) and Less Occluded (LO) subsets with grid loss (G) and hinge loss (H).}
    \label{tbl:occlusion-robustness}
    \end{minipage}
    \hfill
    \begin{minipage}[t]{0.35\textwidth}
    \centering
    \begin{tabular}{l|r}
        \hline
        Method & Correlation \\
        \hline
        Grid loss  & \textbf{225.96} \\
        Hinge loss  & 22500.25 \\
        \hline
    \end{tabular}
    \caption{Grid loss reduces correlation in feature maps.}
    \label{tbl:eval-correlation}
    \end{minipage}
\end{table}

%
%
%

\subsection{Effect on Correlation of Features}
\label{sec:grid-loss-overfitting}

With grid loss we train several classifiers operating on spatially independent image
parts simultaneously. 
During training \acp{CNN} develop discriminative features which are suitable
to classify an image. By dividing the input image into several parts with different appearance,
the \ac{CNN} has to learn features suitable to classify each of these face parts individually.

Since parts which are located on the mouth-region of a face do not contain
e.g. an eye, the \ac{CNN} has to develop features to detect a mouth for this
specific part detector. In contrast to that, with standard loss functions the
\ac{CNN} operates on the full detection window. To classify a given sample as
positive, a \ac{CNN} might solve this classification problem by just learning features which e.g. detect eyes. 
Hence, by operating on the full detection window, only a smaller set of mid-level features is required
compared to \acp{CNN} trained on both, the full detection window and sub-parts.

Therefore, with our method, we encourage \acp{CNN} to learn more diverse
features. More diverse features result in less correlated feature activations,
since for a given sample different feature channels should be active for
different mid-level features.  To measure this, we train a \ac{CNN} with and
without grid loss.  For all spatial coordinates of the last $12 \times 12$
convolution layer, we compute a $128 \times 128$ dimensional normalized
correlation matrix. We sum the absolute values of the off-diagonal elements of the correlation matrices. A
higher number indicates more correlated features and is less desirable. As we
see in Table \ref{tbl:eval-correlation} our grid loss detector learns
significantly less correlated features.




\subsection{Training Set Size}
\label{sec:eval-grid-loss-training-size}
Regularization methods should improve performance of machine learning
methods especially when the available training data set is small. The performance gap between a method
without regularization to a method with regularization should increase with a smaller amount of
training data.
To test the effectiveness of our grid loss as regularization method,
we subsample the positive training samples by a factor of 0.75 - 0.01 and compare the performance
to a standard \ac{CNN} trained on hinge loss, a \ac{CNN} trained with hinge loss on both the output and hidden layers \cite{Szegedy2014, lee2014deeply}, and a \ac{CNN} where we 
apply grid loss on both hidden layers and the output layer.
To assess the performance of each model, we compare the true positive rate at a
false positive count of 284 ($\approx$~0.1~\ac{FPPI}). In
Table~\ref{tbl:training-size-comparison} we see that our grid loss indeed acts
as a regularizer. The performance gap between our method and standard
\acp{CNN} increases from 3.2\% to 10.2\% as the training set gets smaller. Further, we observe that grid loss
benefits from the method of \cite{Szegedy2014, lee2014deeply}, since by applying grid loss 
on top of the hidden layers, the performance gap increases even more. 

\begin{table}[h]
    \centering
    \begin{subfigure}{0.30\textwidth}

    \resizebox{1.0\textwidth}{!}{
        \beginpgfgraphicnamed{fddb-training-set}
        \input{images/fddb-training-set.pgf}
        \endpgfgraphicnamed
    }
    \end{subfigure}
    \begin{subfigure}{0.60\textwidth}
    \small
    \begin{centering}
    \small
    \begin{tabular}{l|ccccccc}
    \hline
    M & 1.00 & 0.75 & 0.50 & 0.25 & 0.10 & 0.05 & 0.01 \\
    \hline
    G-h & \textbf{0.863} & \textbf{0.858} & \textbf{0.856} & \textbf{0.848} & \textbf{0.841} & \textbf{0.833} & \textbf{0.802} \\
    G   & \underline{0.851} & \underline{0.849} & \underline{0.848} & \underline{0.844} & \underline{0.835} & \underline{0.812} & \underline{0.802} \\
    H-h & 0.834 & 0.817 & 0.813 & 0.801 & 0.786 & 0.769 & 0.730 \\
    H   & 0.819 & 0.799 & 0.795 & 0.770 & 0.761 & 0.747 & 0.700 \\
    \hline

    \end{tabular}
    \end{centering}
\end{subfigure}

\caption{
    Impact of training on a sub-set (i.e. 0.75 - 0.01) of the positive training set on \ac{FDDB} at 0.1 \ac{FPPI} using the hinge loss (H), hinge loss on hidden layers (H-h) and our grid loss (G) and grid loss on hidden layers (G-h).}
\label{tbl:training-size-comparison}
\end{table}

\subsection{Ellipse Regressor and Layer Skipping} 
\label{sec:eval-ellipse-regressor}
We compare the impact of an ellipse regressor trained on the 
PASCAL overlap criterion with a regressor trained on the \ac{SSE} loss. We evaluate
the impact on the FDDB dataset using the continuous evaluation protocol~\cite{fddbTech}, 
which weighs matches of ground truth and prediction with their soft PASCAL overlap score. 
In Table~\ref{tbl:ellipse-eval} we see that minimizing the numerical
overlap performs barely better than minimizing the \ac{SSE} loss in the parameter space (i.e. 0.1\% to 0.2\%).
We hypothesize that this is caused by inconsistent annotations in our training set. 

Further, we compare our model with and without an ellipse regressor using different image pyramid sizes.
We evaluate the performance on the FDDB dataset under the discrete evaluation protocol.
In Table~\ref{tbl:ellipse-eval-with-and-without-regressor} we see that regressing ellipses 
improves the true positive rate by about 1\%. But more importantly, using a regressor to refine the face positions allows
us to use fewer intermediate scales in our image pyramid without significant loss in accuracy. 
This greatly improves runtime performance of our detector by a factor of 3-4 (see Sec.~\ref{sec:computational-efficiency}).


\begin{table}[h]


    \centering
    \begin{subfigure}{0.5\textwidth}
        \centering
        \small
        \begin{tabular}{l|ccccc}
            \hline
            Method  & 50 FP  & 100 FP  & 284 FP  & 500 FP  & 1000 FP \\
            \hline
            NUM (D) & \textbf{0.680} & \textbf{0.690} & \textbf{0.702} & \textbf{0.708} & \textbf{0.714} \\
            SSE (D) & 0.679 & 0.688 & 0.700 & 0.706 & 0.713 \\
            \hline
        \end{tabular}
    \end{subfigure}

    \caption{Continuous evaluation of the two proposed ellipse loss functions: Numerical
    PASCAL VOC overlap (NUM) and \ac{SSE} on FDDB.}
    \label{tbl:ellipse-eval}
\end{table}

\begin{table}[h]

    \centering
    \begin{subfigure}{0.4\textwidth} 
    \resizebox{1.0\textwidth}{!}{
        \beginpgfgraphicnamed{fddb-ellipse-regressor-comparison-discrete}
        \input{images/fddb-ellipse-regressor-comparison-discrete.pgf}
        \endpgfgraphicnamed
    }
\end{subfigure}
\begin{subfigure}{0.5\textwidth}
    \centering
    \small
    \begin{tabular}{l|cccc}
    \hline
    Method  & 50 FP  & 100 FP  & 284 FP  & 500 FP  \\
    \hline
    NUM (D) & \underline{0.843} & \textbf{0.857} & \textbf{0.872} & \textbf{0.879}    \\
    NUM (S) & 0.835             & 0.851          & 0.867          & 0.874             \\
    SSE (D) & \textbf{0.844}    & \textbf{0.857} & \textbf{0.872} & \underline{0.878} \\
    SSE (S) & 0.835             & 0.848          & 0.866          & 0.873             \\
    w/o (D) & 0.815             & 0.838          & 0.863          & 0.871             \\
    \hline
    \end{tabular}
\end{subfigure}

\caption{Effect of numerical loss (NUM), \ac{SSE} loss (SSE) and no ellipse
regressor (w/o) applied densely (D) on all pyramid levels or skipping (S)
layers on FDDB.}
\label{tbl:ellipse-eval-with-and-without-regressor}

\end{table}

\subsection{Building a Highly Accurate Detector}
\label{sec:large-network}

Grid loss can also be used to improve the detection performance of deeper networks, yielding highly accurate detections. 
To this end, following
\cite{simonyan2014very}, we replace each $5\times5$ convolution layer with two
$3\times3$ layers, doubling the number of layers from $2$ to $4$. After the first convolution layer 
we apply \ac{LCN}.
Further, we increase the number of convolution filters in our layers to $64$, $256$, $512$
and $512$, respectively. We denote this detector \emph{Big} in the following experiments.

\subsection{Comparison to the State-of-the-Art}
\label{sec:comparison-state-of-the-art}
\begin{figure}[h]
    \centering
    \resizebox{0.42\textwidth}{!}{
    \beginpgfgraphicnamed{AFW-final}
    \input{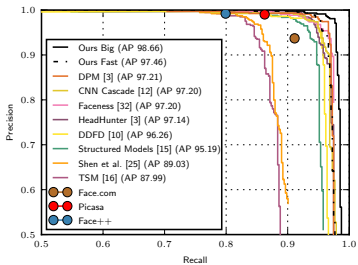}
    \endpgfgraphicnamed
    }
\caption{Our method outperforms state-of-the-art methods on AFW~\cite{ZhuRCVPR2012}.}
\label{fig:afw-eval}
\end{figure}

%
We compare our detector to the state-of-the-art on the FDDB dataset~\cite{fddbTech}, the 
AFW dataset~\cite{ZhuRCVPR2012} and PASCAL Faces dataset~\cite{Yan2014790}, see
Figs.~\ref{fig:fddb-eval-discrete},~\ref{fig:pascal-eval} and~\ref{fig:afw-eval}. For evaluation on 
AFW and PASCAL Faces we use the evaluation toolbox provided by \cite{Mathias2014Eccv}.
For evaluation on FDDB we use the original evaluation tool provided by~\cite{fddbTech}.
We report the accuracy of our small fast model and our large model. 
On FDDB our fast network combined with our regressor retrieves 86.7\% of all
faces at a false positive count of 284, which corresponds to about 0.1 \ac{FPPI}
on this dataset. With our larger model we can improve the true positive rate to 89.4\% at 0.1 \ac{FPPI}, outperforming the state-of-the-art by 0.7\%. 
In our supplementary material we show that when we combine AlexNet with our method, we can increase
the true positive rate to 90.1\%.
On PASCAL Faces and AFW we outperform the state-of-the-art by $1.38\%$ and $1.45\%$ Average Precision respectively.

\begin{figure}[h]
\centering
\begin{minipage}{0.42\textwidth}
    \resizebox{\textwidth}{!}{
        \beginpgfgraphicnamed{fddb-discrete-evaluation}
        \input{images/fddb-discrete-evaluation.pgf}
        \endpgfgraphicnamed
    }
    \caption{Discrete evaluation on the FDDB~\cite{fddbTech} dataset.}
\label{fig:fddb-eval-discrete}
\end{minipage}
\hspace{0.1cm}
\begin{minipage}{0.42\textwidth}
    \resizebox{\textwidth}{!}{
        \beginpgfgraphicnamed{PASCAL-final}
        \input{images/PASCAL_final.pgf}
        \endpgfgraphicnamed
    }
    \caption{Evaluation on the PASCAL Faces~\cite{Yan2014790} dataset.}
    \label{fig:pascal-eval}
\end{minipage}
\end{figure}

\subsection{Computational Efficiency}
\label{sec:computational-efficiency}
%
We implemented our method with Theano~\cite{Bastien-Theano-2012} and Python and
ran our experiments on a desktop machine with a NVIDIA GTX 770 and a
3.20 GHz Intel Core i5 CPU.
Our small dense model needs about 200 ms (GPU) to run on images with a size of $640 \times 480$ pixels.
With skipping intermediate scales our network runs in about 50 ms (GPU) on the same computer 
using non-optimized Python code. On the CPU our small network runs in about 170 ms with layer skipping, achieving competitive 
runtime performance compared to fast tree based methods, e.g. \cite{Mathias2014Eccv, yang2014aggregate}, while outperforming them in accuracy. 
Note that we do not rely on speedup techniques such as
image patchwork~\cite{dubout2012exact, girshick2015dpdpm}, decomposing convolution filters into
separable kernels~\cite{jaderberg2014speeding, zhang2015}, or cascades~\cite{CNNCascade2015}.
Combining our method with these approaches can improve the runtime performance even more.

\section{Conclusion}
%
We presented a novel loss layer named grid loss, which
improves the detection accuracy compared to regular loss layers
by dividing the last convolution layer into several part detectors. This results 
in a detector which is more robust to occlusions compared to standard \acp{CNN}, since each detector
is encouraged to be discriminative on its own.
Further, in our evaluation we observe that \acp{CNN} trained with grid loss develop less correlated
features and that grid loss reduces overfitting.
Our method does not add any additional overhead during runtime.
We
evaluated our detector on face detection tasks and showed that we outperform
competing methods on FDDB, PASCAL Faces and AFW.
The fast version of our method runs at 20 FPS on standard desktop hardware without relying on recently proposed speedup mechanisms, while achieving
competitive performance to state-of-the-art methods. Our accurate model outperforms state-of-the-art methods on public 
datasets while using a smaller amount of parameters.
Finally, our method is complementary to other proposed methods, such as
the \ac{CNN} cascade~\cite{CNNCascade2015} and can improve the discriminativeness
of their feature maps. 

\subsubsection*{Acknowledgements} 
This work was supported by the Austrian Research Promotion Agency (FFG) project DIANGO (840824).

{
\bibliographystyle{splncs}
\bibliography{abbreviation_short,egbib}
}
\clearpage

\end{document}